\documentclass[10pt]{article}
\usepackage{amsfonts}
\usepackage{amsmath}
\usepackage{amssymb}
\usepackage{amsthm}
\usepackage{mathrsfs}
\usepackage{graphicx}
\usepackage{tabularx}

\usepackage{algorithmic} 

\def\R{\mathbb{R}}

\newtheorem{example}{Example}[section] 
\newtheorem{remark}{Remark}[section]
\newtheorem{algorithm}{Algorithm}[section]

\usepackage{amsthm,amsmath,amssymb}

\def\R{\mathbb{R}}

\def\sign{\texttt{sign}}
\def\eig{\texttt{eig}}
\def\diag{\texttt{diag}}

\title {\Large{\bf  Relaxed 2-D Principal Component Analysis by $L_p$ Norm for Face Recognition
}}

\author{Xiao Chen$^1$
Zhi-Gang Jia$^1$\thanks{Corresponding author.  E-mail: zhgjia@jsnu.edu.cn},
Yunfeng Cai$^2$
Mei-Xiang Zhao$^1$,
 \\
$1.$ School of Mathematics and Statistics  \&  Jiangsu Key Laboratory \\ of Education Big Data Science and
Engineering, \\ Jiangsu Normal University,
Xuzhou 221116, China\\
$2.$ Big Data Lab (BDL-US), Baidu Research National Engineering\\
 Laboratory for Deep Learning Technology and Applications, Beijing 100193, China
}

\date{}

\begin{document}
\maketitle

\begin{abstract}
A relaxed two dimensional  principal component analysis (R2DPCA) approach is proposed for face recognition. Different to  the  2DPCA,  2DPCA-$L_1$ and G2DPCA, the R2DPCA utilizes the label information (if known) of training samples to calculate a relaxation vector and presents a weight to each subset of training data. A  new relaxed scatter matrix is defined and  the computed  projection axes are able to increase the accuracy of  face recognition. The optimal $L_p$-norms are selected in a reasonable  range.   Numerical experiments on practical face databased indicate that the R2DPCA has high generalization ability and  can achieve a higher recognition rate  than state-of-the-art methods.
\end{abstract}


\section{Introduction}
\noindent
The principal component analysis (PCA) \cite{Jolliffe04,TP91}, has become one of the most powerful  approaches  of face recognition \cite{siki87,kisi90,tupe91, zhya99,pent00}. Recently, many robust PCA (RPCA) algorithms are proposed with improving the quadratic formulation, which renders PCA vulnerable to noises, into  $L_1$-norm on the objection function, e.g., $L_1$-PCA \cite{keka05}, $R_1$-PCA \cite{dzhz06}, and PCA-$L_1$ \cite{kwak08}. Meanwhile, sparsity is also introduced into PCA algorithms, resulting in a series of sparse PCA (SPCA) algorithms \cite{zht06,agjl07,shhu08,wth09}. A newly proposed robust SPCA (RSPCA) \cite{mzx12} further applies $L_1$-norm both in objective and constraint functions of PCA, inheriting the merits of robustness and sparsity. Observing that $L_2$-, $L_1$-, and $L_0$-norms are all special $L_p$-norm,  it is natural to impose $L_p$-norm on the objection or/and constraint functions, straightforwardly; see  PCA-$L_p$ \cite{kwak14} and generalized PCA (GPCA) \cite{lxzzl13} for instance.

To preserve the spatial structure of face images, two dimensional PCA (2DPCA), proposed by Yang et al. \cite{yzfy04},   represents face images with two dimensional matrices rather than one dimensional vectors.  The computational problems bases on 2DPCA are of much smaller scale than those based on traditional PCA, and the difficulties caused by low rank are also  avoided.
This image-as-matrix method offers insights for improving above  RSPCA,  PCA-$L_p$, GPCA,  etc. As typical examples, the $L_1$-norm-based 2DPCA (2DPCA-$L_1$) \cite{lpy10} and 2DPCA-$L_1$ with sparsity (2DPCA$L_1$-S) \cite{whwj13} are improvements of PCA-$L_1$ and RSPCA, respectively, and the generalized 2DPCA (G2DPCA) \cite{wangj16} imposes $L_p$-norm on both objective and constraint functions of 2DPCA. 
Recently, the quaternion 2DPCA is proposed in \cite{jlz17} and applied to color face recognition, where the red, green and blue channels of a color image is encoded as three imaginary parts of a pure  quaternion matrix.   To arm the quaternion 2DPCA with the generalization ability,  Zhao, Jia and Gong \cite{zjg17} proposed the sample-relaxed quaternion 2DPCA with applying  the label information (if known) of training samples. The structure-preserving algorithms of quaternion eigenvalue decomposition and singular value decomposition can be found in \cite{jwl13,mjb18,jmz17,jwzc18,jcz09,jww10,zhji14, jns18b}.  More applications of the quaternion representation and structure-preserving methods  to  color image processing can be found in \cite{jns18a} and \cite{jnw18}.

Both PCA and 2DPCA are unsupervised methods and omit the potential or known  label information of samples.  They are often applied to the training set and thus the computed projections will maximize the scatter of projected training samples.  That means the scatter of projected testing samples are not surely optimal, and certainly, so are the whole (training and testing) projected samples.  Inspired by this observation, we proposed a new relaxation two-dimensional principal component analysis (R2DPCA) in this paper.
R2DPCA sufficiently utilizes the labels (if known) of training samples, and can enhance  the total scatter of whole projected samples. 
 This approach is a generalization of G2DPCA \cite{wangj16}, and will reduce to G2DPCA if the label information is unknown or unused.

The rest of this paper is organized as follows. In Section \ref{s:face}, we recall robust and sparse 2DPCA algorithms.  In Section \ref{s:r2dpca}, we present a new relaxed two dimensional  principal component analysis (R2DPCA) approach for face recognition.  In Section \ref{s:exs}, we compare the R2DPCA with the state-to-the-art approaches, and indicate the efficiencies of the R2DPCA . In Section \ref{s:conclusion}, we sum up the contribution of this paper.

\section{Robust and sparse 2DPCA algorithms }\label{s:face}
In this section, we recall  2DPCA, 2DPCA-$L_1$, 2DPCA$L_1$-S, and G2DPCA algorithms in the form of computing the first projection vector. In fact, after obtaining first $j$ projection vectors $\textbf{W}=[\textbf{w}_{1},\textbf{w}_{2},\ldots,\textbf{w}_{j}]$, the $(j+1)$-th projection vector $\textbf{w}_{j+1}$ can be calculated similarly on deflated samples \cite{Mackey08}:
\begin{equation}\label{11}
  \textbf{X}_{i}^{deflated}=\textbf{X}_{i}(\textbf{I}-\textbf{W}\textbf{W}^{T}), i=1,2,\ldots,n.
\end{equation}

\subsection{2DPCA}

Suppose  that  there are $n$ training images samples $ \textbf{X}_{1},\textbf{X}_{2},\ldots,\textbf{X}_{n} \in\mathbb{R}^{h\times w}$, where $h$ and $w$ denote the height and width of images, respectively.  We assume that these samples are mean-centered, i.e., $\frac{1}{n}\sum^{n}\nolimits_{i=1}\textbf{X}_{i}=0$; otherwise, we will replace $\textbf{X}_i$ by $\textbf{X}_i-\frac{1}{n}\sum^{n}\nolimits_{i=1}\textbf{X}_i$.

 2DPCA \cite{yzfy04}
finds its first projection vector $\textbf{w}\in\mathbb{R}^{w}$ by solving the optimization problem with equality constraints:
\begin{equation}\label{1}
  \max\limits_{\textbf{w}\in\mathbb{R}^{w}}\sum\limits^{n}_{i=1}\|\textbf{X}_{i}\textbf{w}\|_{2}^{2}, \  s.t. \|\textbf{w}\|_{2}^{2}=1.
\end{equation}
The projection vector $\textbf{w}$ could be calculated by the iterative algorithm:
\begin{subequations}\label{e:pmethod_2}
\begin{align}
         &\textbf{v}^{k}=\sum\limits^{n}_{i=1}\textbf{X}_{i}^{T}[|\textbf{X}_{i}\textbf{w}^{k}|\circ \sign(\textbf{X}_{i}\textbf{w}^{k})],\\
         &\textbf{u}^{k}=|\textbf{v}^{k}|\circ \sign(\textbf{v}^{k}),\\
       &\textbf{w}^{k+1}=\frac{\textbf{u}^{k}}{\|\textbf{u}^{k}\|_{2}},
     \end{align}
 \end{subequations}
 where  \sign $(\cdot)$ denotes the sign function.
The projection vector $\textbf{w}$ can also be obtained by calculating the eigen decomposition of a covariance matrix and selecting the eigenvector corresponding to the largest eigenvalue. See Remark \ref{remark1} for more details.

\subsection{2DPCA-$L_1$}
 2DPCA-$L_1$ \cite{lpy10} finds its first projection vector $\textbf{w}\in\mathbb{R}^{w}$ by solving  the optimization problem with equality constraints:
\begin{equation}\label{2}
  \max\limits_{\textbf{w}\in\mathbb{R}^{w}}\sum\limits^{n}_{i=1}\|\textbf{X}_{i}\textbf{w}\|_{1}, \  s.t. \|\textbf{w}\|_{2}^{2}=1.
\end{equation}
The projection vector $\textbf{w}$ could be calculated by the iterative algorithm:
 \begin{subequations}
 \begin{align}
         &\textbf{v}^{k}=\sum\limits^{n}_{i=1}\textbf{X}_{i}^{T}\sign(\textbf{X}_{i}\textbf{w}^{k}),\\
         &\textbf{w}^{k+1}=\frac{\textbf{v}^{k}}{\|\textbf{v}^{k}\|_{2}},
 \end{align}
 \end{subequations}
where   $\textbf{w}^{k}$ is the projection vector at the $k$-th step. Notice that 2DPCA-$L_1$  could be formulated by replacing the $L_2$-norm in  objective function of 2DPCA with $L_1$-norm.

\subsection{2DPCA$L_1$-S}
 2DPCA$L_1$-S \cite{whwj13}
finds its first projection vector $\textbf{w}\in\mathbb{R}^{w}$ by solving the optimization problem with equality and inequality constraints:
\begin{equation}\label{5}
  \max\limits_{\textbf{w}\in\mathbb{R}^{w}}\sum\limits^{n}_{i=1}\|\textbf{X}_{i}\textbf{w}\|_{1}, \  s.t. \|\textbf{w}\|_{1}\leq c, \|\textbf{w}\|_{2}^{2}=1,
\end{equation}
where $c$ is a positive constant.
The projection vector $\textbf{w}$ could be calculated by the iterative algorithm:
\begin{subequations}
\begin{align}
        &\textbf{v}^{k}=\sum\limits^{n}_{i=1}\textbf{X}_{i}^{T}\sign(\textbf{X}_{i}\textbf{w}^{k}),\\
  \label{e:lam}       & u_{i}^{k}=v_{i}^{k}\frac{|w_{i}^{k}|}{\lambda+|w_{i}^{k}|}, \ i=1,2,\ldots,w,\\
       &\textbf{w}^{k+1}=\frac{\textbf{u}^{k}}{\|\textbf{u}^{k}\|_{2}},
\end{align}
 \end{subequations}
where $u_{i}^{k}$, $v_{i}^{k}$, and $w_{i}^{k}$ are the $i$th elements of vectors $\textbf{u}^{k},\textbf{v}^{k},$ and $\textbf{w}^{k}\in\mathbb{R}^{w}$, respectively.
In equation \eqref{e:lam}, $\lambda$ is a positive scalar which serves as a tuning parameter. When $\lambda$ is set to be zero, 2DPCA$L_1$-S reduces to 2DPCA-$L_1$.
Notice that 2DPCA$L_1$-S could be formulated by  imposing $L_1$-norm on  objective and constraint functions of 2DPCA.
\subsection{G2DPCA}
G2DPCA \cite{wangj16} finds its first projection vector $\textbf{w}\in\mathbb{R}^{w}$ by solving the optimization problem with equality constraints:
\begin{equation}\label{9}
  \max\limits_{\textbf{w}\in\mathbb{R}^{w}}\sum\limits^{n}_{i=1}\|\textbf{X}_{i}\textbf{w}\|_{s}^{s}, \  s.t. \|\textbf{w}\|_{p}^{p}=1
\end{equation}
where $s\geq1$ and $p>0$.  The projection vector $\textbf{w}$ can be updated in two different ways,  depending on the value $p$.
\begin{itemize}
\item[] {\bf Case 1:} If $p\geq1$,
\begin{subequations}\label{e:pmethod}
\begin{align}
         &\textbf{v}^{k}=\sum\limits^{n}_{i=1}\textbf{X}_{i}^{T}[|\textbf{X}_{i}\textbf{w}^{k}|^{s-1}\circ \sign(\textbf{X}_{i}\textbf{w}^{k})],\\
         &\textbf{u}^{k}=|\textbf{v}^{k}|^{q-1}\circ \sign(\textbf{v}^{k}),\\
       &\textbf{w}^{k+1}=\frac{\textbf{u}^{k}}{\|\textbf{u}^{k}\|_{p}}.
     \end{align}
 \end{subequations}
where $q$ satisfies $1/p+1/q=1$, $\circ$ denotes the Hadamard product, i.e., the element-wise product between two vectors. Especially, if  $p=1$, $\textbf{w}^{k+1}$ can be computed by
\begin{subequations}
     \begin{align}
        &j=\arg \max\nolimits_{i\in [1,w]}|v_{i}^{k}|,\\
        &w_{i}^{k+1}=\left\{
\begin{aligned}
\sign(v_{j}^{k}), \ i=j,\\
0, \ i\neq j,\\
\end{aligned}
\right.
     \end{align}
 \end{subequations}
wherein $v_{i}^{k}$ is the $i$-th value of $\textbf{v}^{k}$;
if $p=\infty$, $\textbf{w}^{k+1}$ can be computed by
\begin{equation}\label{15}
  \textbf{w}^{k+1}=\sign(\textbf{v}^{k}).
\end{equation}
\item[] {\bf Case 2:} If  $0<p<1$,
\begin{subequations}\label{e:pmethod1}
     \begin{align}
        &\textbf{v}^{k}=\sum\limits^{n}_{i=1}\textbf{X}_{i}^{T}[|\textbf{X}_{i}\textbf{w}^{k}|^{s-1}\circ \sign(\textbf{X}_{i}\textbf{w}^{k})],\\
     &\textbf{u}^{k}=|\textbf{w}^{k}|^{2-p}\circ \textbf{v}^{k},\\
       &\textbf{w}^{k+1}=\frac{\textbf{u}^{k}}{\|\textbf{u}^{k}\|_{p}}.
     \end{align}
 \end{subequations}
 \end{itemize}
Notice that G2DPCA  could be formulated by generalizing $L_2$-norm  in objective and constraint functions of 2DPCA to $L_s$-norm and $L_p$-norm, respectively.

\begin{remark}\label{remark0} When $s=p=2$, the projection method \eqref{e:pmethod} reduces to \eqref{e:pmethod_2}.
\end{remark}

\begin{remark}\label{remark1}
By the eigenvalue decomposition method,  2DPCA can  select a set of projection axes $\{\textbf{w}_{1},\textbf{w}_{2},\cdots,\textbf{w}_{r}\}$  in one step, without selecting only one optimal projection axis each step.
These projection axes are chosen as eigenvectors of a covariant matrix corresponding to first $r$ largest eigenvalues:
 \begin{subequations}\label{e:eigmethod}
     \begin{align}
&\textbf{G}_{t}=\frac{1}{n}\sum\limits^{n}_{i=1}\textbf{X}_{i}^{T}\textbf{X}_{i},\\
&[\textbf{W},\textbf{D}]=\eig(\textbf{G}_{t}),
     \end{align}
 \end{subequations}
where $\textbf{G}_{t}$ is the covariance matrix of training samples,
$\textbf{W}=[\textbf{w}_{1}, \textbf{w}_{2}, \cdots, \textbf{w}_{r}]\in R^{w\times r}$ is a matrix with unitary column vectors,
$\textbf{D}$ is a diagonal matrix consists of first $r$ largest eigenvalues.
Since $G_t$ is symmetric and positive semi-definite,  the diagonal elements of $D$ are nonnegative and  the projection axes
$$ \{\textbf{w}_{1},\textbf{w}_{2},\cdots,\textbf{w}_{r}\}=\arg \max \sum\limits^{n}_{i=1} \textbf{w}_{i}^{T} \textbf{G}_{t} \textbf{w}_{i}, $$
are orthogonal to each other, i.e.,
\begin{equation}
\textbf{w}_i^T\textbf{w}_j=\left\{
\begin{array}{l}  1,~ i=j,\\  0,~i\neq j, \end{array}
~s,t=1,\cdots,r.
\right.
\end{equation}
\end{remark}

\begin{remark}
We sum the procedures of above four methods in Table \ref{compare}.  Their relationship is also indicated  in Fig \ref{relationship}.
\begin{table*}\label{compare}
\centering
\small{\caption{Procedures }{\label{compare}}}
\begin{tabular}{c|l|lc}
 \hline
  {\bf Algorithm}~~~~&  ~~~~{\bf 2DPCA}~~~~&  ~~~~{\bf 2DPCA-$L_1$}~~~~  &\\ \hline
{\bf Procedure} &                $[\textbf{W},\textbf{D}]=\eig(\frac{1}{n}\sum\limits^{n}_{i=1}\textbf{X}_{i}^{T}\textbf{X}_{i})$ &           $\textbf{v}^{k}=\sum\limits^{n}_{i=1}\textbf{X}_{i}^{T}\sign(\textbf{X}_{i}\textbf{w}^{k})$ \\
$\ $&     
&        $\textbf{w}^{k+1}=\frac{\textbf{v}^{k}}{\|\textbf{v}^{k}\|_{2}}$  \\\hline\hline
{\bf Algorithm}&                       ~~~~{\bf 2DPCA$L_1$-S}~~~~&  ~~~~{\bf G2DPCA}~~~~  &   \\ \hline
{\bf Procedure}&                      $\textbf{v}^{k}=\sum\limits^{n}_{i=1}\textbf{X}_{i}^{T}\sign(\textbf{X}_{i}\textbf{w}^{k}) $ &    $\textbf{v}^{k}=\sum\limits^{n}_{i=1}\textbf{X}_{i}^{T}[|\textbf{X}_{i}\textbf{w}^k|^{s-1}\circ \sign(\textbf{X}_i\textbf{w}^k)]$    \\
$\ $&          $u_{i}^{k}=v_{i}^{k}\frac{|w_{i}^{k}|}{\lambda+|w_{i}^{k}|}, \ i=1,2,\ldots,w$ &     \textbf{if} $ p\ge 1$  ~~  \textbf{run}  \eqref{e:pmethod}      \\
$\ $&         $\textbf{w}^{k+1}=\frac{\textbf{u}^{k}}{\|\textbf{u}^{k}\|_{2}} $ &    \textbf{if} $0<p< 1$  ~~  \textbf{run}  \eqref{e:pmethod1}    \\
\hline
\end{tabular}
\small
\end{table*}
\end{remark}
\section{The relaxed 2DPCA by $L_p$-norm}\label{s:r2dpca}
\noindent
In this section,  we introduce a  relaxed two-dimensional principal component analysis (R2DPCA) method by $L_p$-norm.
R2DPCA  includes three parts:  relaxation vector generation, objective function relaxation, and projection relaxation.
\subsection{Relaxation vector}\label{ss:rv}

Suppose that training samples $\textbf{X}_1, \textbf{X}_2, ..., \textbf{X}_n \in\R^{h\times w}$ can be partitioned into $m$ classes and each class contains $n_j$  samples:
$$\textbf{X}_1^1, \cdots, \textbf{X}_{n_1}^1\ | \ \textbf{X}_1^2, \cdots, \textbf{X}_{n_2}^2\ |\  \cdots \ |\  \textbf{X}_1^m, \cdots, \textbf{X}_{n_m}^m,$$
where $\textbf{X}_i^j$ denotes the $i$-th sample of the $j$-th class, $i=1,\ldots, n_j$, $j=1,\ldots, m$.
Define the mean of training samples from the $j$-th class as
 $$\textbf{M}_j=\frac{1}{n_{j}}\sum \limits_{i=1}^{n_{j}}\textbf{X}_i^{j}\in\R^{h\times w},$$
 and  
  the $j$-th within-class covariance matrix of the training set  as
 \begin{equation}\label{e:withincovariancematrix}
\textbf{C}_j=\frac{1}{n_{j}}\sum\limits_{i=1}^{n_{j}}(\textbf{X}_i^j-\textbf{M}_j)^T(\textbf{X}_i^j-\textbf{M}_j)\in\R^{w\times w},
\end{equation}
where $j=1,\ldots,m$, $\sum_{j=1}^{m}n_j=n$ and $i=1,\ldots,n_j$.

The within-class covariance matrix $\textbf{C}_j$ is a symmetric and positive semi-definite  matrix. 
Its maximal eigenvalue, denoted by $\lambda_{\max}(\textbf{C}_j)$, represents the variance of training samples $\textbf{X}_1^j, \ldots, \textbf{X}_{n_j}^j$
 in the principal component. In general, the larger $\lambda_{\rm max}(\textbf{C}_j)$ is,  the better scattered of the training samples of $j$-th class are.  A very small $\lambda_{\rm max}(\textbf{C}_j)$ indicates that  $\textbf{X}_1^j, \ldots, \textbf{X}_{n_j}^j$ are not well scattered samples to represent the $j$-th class.
Extremely,  if $\lambda_{\max}(\textbf{C}_j)=0$ then all of training samples from the $j$-th class are same, and then the contribution of the $j$-th class  to the covariance matrix of training set should be controlled by a small factor. To this aim, we define a {\it relaxation vector} of training classes,
 \begin{equation}\label{e:v}
 \textbf{v}=[v_1,\cdots,v_m]^T\in\mathbb{R}^m,   
 \end{equation}
 where
 \begin{equation}\label{rfactor}
   v_j=\frac{f(\lambda_{\rm max}(\textbf{C}_j)) }{\sum_{i=1}^{m}f(\lambda_{\rm max}(\textbf{C}_i)) },
 \end{equation}
 is a relaxation factor of the $j$-th class with a function, $f:~\R\rightarrow \R^{+}$.
A relaxation factor of each training sample of $j$-th class is defined as $v_j/n_j$. If each training class has only one  sample, i.e.,  $n_1=\cdots=n_m=1$,  then all within-class covariance matrices are zero matrix and
$\lambda_{\max}(\textbf{C}_1)=\cdots=\lambda_{\max}(\textbf{C}_m)=0.$
In this case, the relaxation factor of each class is  same ($v_j=1/m$), and so is the factor of each training sample of $j$-th class.

We sum above steps of computing the relaxation vector of training set in Algorithm \ref{a:rv}.
\begin{algorithm}
{\bf Relaxation vector generation}
\label{a:rv}
\begin{algorithmic}[section]\small
\STATE {\bf function} $\textbf{v}=\texttt{relaxvec}(\textbf{X}_1, \textbf{X}_2,\cdots, \textbf{X}_n,m,w)$
\FOR {$j=1,2,\cdots,m$}
\STATE $\textbf{C}_j=\texttt{zeros}(w,w);$
\STATE $\textbf{M}_j=\frac{1}{n_j}(\textbf{X}_1^j+\cdots+\textbf{X}_{n_j}^j);$
\FOR{$i=1,2,\cdots,n_j$}
\STATE $\textbf{C}_j=\textbf{C}_j+{(\textbf{X}_i^j-\textbf{M}_j)}'*(\textbf{X}_i^j-\textbf{M}_j);$
\ENDFOR
\STATE $\textbf{C}_j=\textbf{C}_j/n_j;$
\STATE {\bf Compute relaxation vector \textbf{v} defined as in} \eqref{e:v} and \eqref{rfactor};
\ENDFOR
\end{algorithmic}
\end{algorithm}

\subsection{Objective function relaxation}\label{ss:ofr}
Let $\textbf{M}$ denote the mean of training samples, i.e.,
 $$\textbf{M}=\frac{1}{n}\sum\limits^{n}_{i=1}\textbf{X}_i=\frac{1}{n}\sum\limits_{j=1}^{m}\sum\limits_{i=1}^{n_j}\textbf{X}_i^j.$$
With computed relaxation vector $\textbf{v}=$ $[v_1,$ $\cdots, $ $v_m]^T$ in Section \ref{ss:rv},
we define a {\it relaxed criterion} as
\begin{equation}\label{e:gtsc}
J(\textbf{w})=\gamma\textbf{G}+(1-\gamma)\widetilde{\textbf{G}},
\end{equation}
where  $ \gamma\in[0,1]$ is a relaxation parameter,   $\textbf{w}\in\R^w$ is a unit vector under $L_p$ norm, $\textbf{G}:=\sum\limits^{n}_{i=1}\|(\textbf{X}_{i}-\textbf{M})\textbf{w}\|_{s}^{s}$ and $\widetilde{\textbf{G}}:=\sum\limits^{m}_{j=1}\sum\limits^{n_j}_{i=1}\|\frac{v_j}{n_j}(\textbf{X}_i^j-\textbf{M})\textbf{w}\|_{s}^{s}$.
R2DPCA finds its first projection vector $\textbf{w}\in\mathbb{R}^{w}$ by solving the optimization problem with equality  constraints:
\begin{equation}\label{e:r2dcri}
\max\limits_{\textbf{w}\in\R^w} J(\textbf{w}), s.t. \|\textbf{w}\|_p^p=1,
\end{equation}
where the criterion $J(\textbf{w})$ is defined as in \eqref{e:gtsc}.
Notice that \eqref{e:r2dcri} reduces  to \eqref{9} if $\gamma=1$, and thus,  the first projection vector of R2DPCA is the same as that of G2DPCA.
When $\gamma=0$,  \eqref{e:r2dcri} is simplified as
\begin{equation}\label{e:r2dcri_s}
\max\limits_{\textbf{w}\in\R^w}\sum\limits^{m}_{j=1}\sum\limits^{n_j}_{i=1}\|\frac{v_j}{n_j}(\textbf{X}_i^j-\textbf{M})\textbf{w}\|_{s}^{s}, s.t. \|\textbf{w}\|_p^p=1.
\end{equation}
If first $j$ projection vectors $\textbf{W}=[\textbf{w}_{1},\textbf{w}_{2},\ldots,\textbf{w}_{j}]$ have been obtained, the $(j+1)$-th projection vector $\textbf{w}_{j+1}$ can be calculated similarly on the deflated samples, defined as in \eqref{11}.  From each iterative step,
 we also obtain a maximized objective function value corresponding to $\textbf{w}_{j}$,
   $$f_j=\gamma\sum\limits^{n}_{i=1}\|(\textbf{X}_{i}-\textbf{M})^{deflated}\textbf{w}_j\|_{s}^{s}+(1-\gamma)\sum\limits^{m}_{j=1}\sum\limits^{n_j}_{i=1}\|\frac{v_j}{n_j}(\textbf{X}_i^j-\textbf{M})^{deflated}\textbf{w}_j\|_{s}^{s}.$$
 With the relaxed criterion defined in \eqref{e:gtsc},    first $r$ optimal projection vectors of R2DPCA solve the optimal problem with equality constraints:
 \begin{equation}\label{e:r2d4ws}
\begin{array}{l}
\{\textbf{w}_1,\ldots,\textbf{w}_r\}=\arg \max J(\textbf{w})\\ [5pt]
\ {\rm s.t.}\ \
\left\{
\begin{array}{l}  \|\textbf{w}_i\|_p^p=1,\\
 \textbf{w}_i^T\textbf{w}_j=0 ~i\neq j,
 \end{array}
~i,j=1,\cdots,r.
\right.
\end{array}
\end{equation}
We propose  Algorithm \ref{a:r2dpca} to compute first $r$ optimal projection vectors,  $\textbf{W}=[\textbf{w}_1, \cdots, \textbf{w}_r]$, and corresponding optimal objective function values,  $\textbf{D}={\rm diag}(f_1,\ldots, f_r)$.

\begin{algorithm}
{\bf R2DPCA}
\label{a:r2dpca}  %
\begin{algorithmic}[section]\small %
\REQUIRE {$\textbf{X}_1, \textbf{X}_2,\cdots, \textbf{X}_n,  s\in [1,\infty),  p\in(0,\infty], r, m, w, \gamma\in[0,1], n_1,\cdots,n_m, tol. $}
\ENSURE $\textbf{W}=[ \textbf{w}_1,\ldots,\textbf{w}_r], \textbf{D}=\diag( f_1,\ldots,f_r).$
\STATE Initialize $\textbf{W}=[ \ ]$,  $\textbf{D}=[ \ ]$. 
\STATE $\textbf{v}$=\texttt{relaxvec}($\textbf{X}_1, \textbf{X}_2,\cdots, \textbf{X}_n, m, w$).
\STATE Homogenize training samples.
\FOR{$t=1,2,\cdots,r$}
\STATE Initialize $k=0$, $\delta=1$, arbitrary $\textbf{w}^0$ with $\parallel\textbf{w}^0\parallel_p=1$.
\STATE $f_0=\gamma\sum\limits^{n}_{i=1}\|\textbf{X}_{i}\textbf{w}^0\|_s^s
+(1-\gamma)\sum\limits^{m}_{j=1}\sum\limits^{n_j}_{j=1}\|\frac{v_j}{n_j}\textbf{X}_{i}^j\textbf{w}^0\|_s^s.$
\WHILE{$\delta> tol$}
\STATE $\textbf{v}^{k}=\gamma\sum\limits^{n}_{i=1}\textbf{X}_{i}^{T}[|\textbf{X}_{i}\textbf{w}^k|^{s-1}\circ \sign(\textbf{X}_i\textbf{w}^k)]
+(1-\gamma)\sum\limits^{m}_{j=1}\sum\limits^{n_j}_{j=1}(\frac{v_j}{n_j}\textbf{X}_{i}^j)^{T}[|\frac{v_j}{n_j}\textbf{X}_{i}^j\textbf{w}^k|^{s-1}\circ \sign(\frac{v_j}{n_j}\textbf{X}_{i}^j\textbf{w}^k)]$.
\STATE \textbf{Case 1:} $0<p<1$
\STATE  \quad $\textbf{u}^k=|\textbf{w}^k|^{2-p}\circ\textbf{v}^k,$
\STATE \quad $\textbf{w}^{k+1}=\frac{\textbf{u}^k}{\parallel\textbf{u}^k\parallel_p}.$
\STATE \textbf{Case 2:} $p=1$
\STATE \quad $j=\arg \max\nolimits_{i\in [1,w]}|v_{i}^{k}|,$
\STATE \quad $ w_{i}^{k+1}=\left\{
\begin{aligned}
\sign(v_{j}^{k}), \ i=j,\\
0, \ i\neq j.\\
\end{aligned}
\right.$
\STATE \textbf{Case 3:} $1<p<\infty$
\STATE \quad$q=p/(p-1),$
\STATE \quad$\textbf{u}^k=|\textbf{v}^k|^{q-1}\circ \sign(\textbf{v}^k),$
\STATE \quad$\textbf{w}^{k+1}=\frac{\textbf{u}^k}{\parallel\textbf{u}^k\parallel_p}.$
\STATE \textbf{Case 4:} $p=\infty$
\STATE \quad$\textbf{w}^{k+1}=\sign(\textbf{v}^k).$
\STATE $f_{k+1}=\gamma\sum\limits^{n}_{i=1}\|\textbf{X}_{i}\textbf{w}^{k+1}\|_{s}^{s}
+(1-\gamma)\sum\limits^{m}_{j=1}\sum\limits^{n_j}_{j=1}\|\frac{v_j}{n_j}\textbf{X}_{i}^j\textbf{w}^{k+1}\|_{s}^{s}.$ \STATE $\delta=|f_{k+1}-f_{k}|/|f_k|.$
\STATE $k\leftarrow k+1.$
\ENDWHILE
\STATE $\textbf{W}\leftarrow [\textbf{W},\textbf{w}^{k}].$
\STATE $\textbf{D}=\diag(\textbf{D},f_k).$
\STATE $\textbf{X}_i=\textbf{X}_i(\textbf{I}-\textbf{W}\textbf{W}^T), i=1,2,\cdots,n.$
\ENDFOR
\end{algorithmic}
\end{algorithm}

\subsection{Projection relaxation} \label{sss:eigenfaces}
 In Section \ref{ss:ofr}, we  obtain $r$ pairs of optimal values and projection vectors: $(f_1, \textbf{w}_1),\ldots, (f_r, \textbf{w}_r)$.
Define  the  {\it feature image} of sample $\textbf{X}_i$ under $\textbf{W}$ as
\begin{equation}\label{e:ps4fs}\textbf{P}_i=(\textbf{X}_i-{\bf{M})}\textbf{W}
 \in\mathbb{R}^{w\times r},\ i=1,\cdots,n.\end{equation}
 Each column of  $\textbf{P}_i$,  $\textbf{y}_j=(\textbf{X}_i-{\bf{M}})\textbf{w}_j$,   is called the {\it principal component (vector)}.

Now we use a nearest neighbour classifier for face recognition. For a given testing sample $\textbf{X}$,  compute its feature image, $\textbf{P}=(\textbf{X}-{\bf{M}})\textbf{W}$.
Find out the nearest training sample $\textbf{X}_i$ $(1\le i\le n)$ whose feature image minimizes
$$\|(\textbf{P}_i-\textbf{P})\textbf{D}\|_2.$$
Such $\textbf{X}_i$ is output as the person to be recognized.

The distance, $\|(\textbf{P}_i-\textbf{P})\textbf{D}\|_2=\|(\textbf{X}_i-\textbf{X})\textbf{W}\textbf{D}\|_2$, is called {\it relaxed distance} between $\textbf{X}_i$ and $\textbf{X}$.
Compared with originally defined distance, such as in \cite{wangj16},
each projection axe $\textbf{w}_j$ is relaxed by $f_j$  in classification process, $j=1,\cdots,r$.

\subsection{Restarted alternating direction search method}\label{ss:radsm}
\noindent
In the R2DPCA approach of face recognition, we need choose optimal $L_s$- and $L_p$-norms  to maximize the recognition or classification rate. The traverse method will cost a huge amount of computational time.  Instead, we present a restarted alternating direction search method of searching optimal values of $s$ and $p$; see Algorithm \ref{a:radsm}.
\begin{algorithm}
{\bf Restarted alternating direction search method}
\label{a:radsm}
\begin{algorithmic}[section]\small
\REQUIRE {A finite range of  $(s,p)$: $\Omega=\{(s_i,p_j)|i,j=1,2,\cdots,N\}$, and  a positive number $\delta$.}
\ENSURE $s, t$.
\STATE \textbf{Step  1.}  Choose a starter  $(s_i^0,p_j^0)\in\Omega$, randomly, and compute the recognition rate, denoted as $\varrho^0(i,j)$.
\STATE \textbf{Step  2.}  With $p_j=p_j^0$, find the maximal recognition rate  in  $\{(s_i,p_j^0)|$ $i=1,2,\cdots,N\}$, denoted as $\varrho^{1/2}(i,j)$, and  denote the maximum point as  $(s_i^1,p_j^0)$.
\STATE \textbf{Step 3.}  With $s_i=s^1_i$, find the maximal recognition rate  in  $\{(s_i^1,p_j)|j=1,2,  \cdots, $ $N\}$,denoted as $\varrho^1(i,j)$,  and  denote the maximum point as  $(s_i^1,p_j^1)$.
\STATE \textbf{Step 4.} If $\varrho^1(i,j)=\varrho^0(i,j)$ and  $(s_i^1,p_j^1)=(s_i^0,p_j^0)$, go to \textbf{Step 5}; otherwise, let $(s_i^0,p_j^0)=(s_i^1,p_j^1)$ be a new starter, and go to \textbf{Step 2}.
\STATE \textbf{Step  5.}  Find  the maximal recognition rate $\varrho^2(i,j)$ in $\{(s_i,p_j)| |s_i-s_i^1|\le \delta, |p_j-p_j^1|\le\delta\}\cap\Omega$, and denote the maximum point as  $(s_i^2,p_j^2)$.
\STATE \textbf{Step  6.}  If $\varrho^2(i,j)\le \varrho^1(i,j)$, output $s=s_i^1$ and $p=p_j^1$; otherwise, let  $(s_i^0,p_j^0)=(s_i^2,p_j^2)$ be a new starter, and go to \textbf{Step 2}.
\end{algorithmic}
\end{algorithm}

\begin{figure}[!h]
  \begin{center}
\includegraphics[height=0.30\textwidth,width=0.45\textwidth]{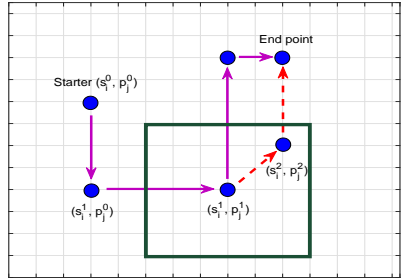}
  \end{center}
  \vskip-15pt \caption{The searching path of restarted alternating direction search method}
\label{path_sec4}
\end{figure}

 If giving a enough large value $\delta$ in Algorithm \ref{a:radsm}, we can surely achieve the maximum value of recognition rate at optimal values $(s_j^2,p_j^2)\in \Omega$.  The selecting process is indicated in  Fig \ref{path_sec4}.

\subsection{Mathematical theory of R2DPCA}
R2DPCA is a generalization of G2DPCA \cite{wangj16}.  As one of PCA-based methods, G2DPCA does not use the labels of data which possibly can impair class discrimination.  To  improve this, R2DPCA utilizes  labels of training samples and variances within class to generate  a relaxation vector,      computes optimal projections maximizing  the relaxes criterion, and thus  enhances the  class discrimination.

The working principle of R2DPCA can be clearly explained through a special case that  $p=s=2$.
The  relaxed criterion with $L_2$-norm is   also called  {\it generalized total scatter criterion},  and has the form:
\begin{equation}\label{e:gtsc2}
J(\textbf{W})={\rm trace}(\textbf{W}^{T}(\gamma\textbf{G}+(1-\gamma)\widetilde{\textbf{G}})\textbf{W})=\sum_{i=1}^{r}\textbf{w}_i^T(\gamma\textbf{G}+(1-\gamma)\widetilde{\textbf{G}})\textbf{w}_i,
\end{equation}
where
\begin{subequations}\label{e:GGt}
   \begin{align}
 \label{e:G}
\textbf{G}&=\frac{1}{n}\sum_{i=1}^{n}(\textbf{X}_i-{\bf{M}})^T(\textbf{X}_i-{\bf{M}}),\\
\label{e:Gt}
\widetilde{\textbf{G}}&=\sum_{j=1}^m\left(\frac{v_j}{n_j}\sum_{i=1}^{n_j}(\textbf{X}_i^j-{\bf{M}}) ^T(\textbf{X}_i^j-{\bf{M}}) \right),
\end{align}
\end{subequations}
where  $\textbf{W}=[\textbf{w}_1,\cdots, \textbf{w}_r]$ has orthogonal columns and each column is unitary under $L_p$-norm,  $v_j$ is the $j$-th element of the relaxation vector $\textbf{v}$.
Here $\sum_{j=1}^m n_j=n$. Recall that ${\bf{M}}\in\mathbb{R}^{h\times w}$ is the mean  of  training samples.
Let $\textbf{W}^{opt}=[\textbf{w}_1^{opt},\cdots, \textbf{w}_r^{opt}]$ be  the optimal projection, where   $\textbf{w}_1^{opt},\cdots, \textbf{w}_r^{opt}$ solve the optimal problem \eqref{e:r2d4ws}. These optimal projection axes are in fact the orthogonal eigenvectors of $\gamma\textbf{G}+(1-\gamma)\widetilde{\textbf{G}}$ corresponding to first $r$ largest eigenvalues. Since  the matrix  $\gamma\textbf{G}+(1-\gamma)\widetilde{\textbf{G}}$ is symmetric and positive semi-definite,   $J(\textbf{W})$ is nonnegative.

R2DPCA with $s=p=2$ can also be seen as applying the relaxation idea to 2DPCA, and thus called relaxed 2DPCA.    Algorithm \ref{a:r2dpca} with $s=2, p=2$  is one method of processing the relaxed 2DPCA.  Another method is applying the eigenvalue decomposition (Algorithm \ref{relaxed_2dpca} ), as shown in Remark \ref{remark1}.
\begin{algorithm}
{\bf relaxed 2DPCA.}
\label{relaxed_2dpca}
\begin{algorithmic}[section]\small
\REQUIRE {$\textbf{X}_1, \textbf{X}_2,\cdots, \textbf{X}_n, \textbf{M}, s=p=2,r, m, w. $}
\ENSURE $\textbf{W}, \textbf{D}$.
\STATE $\textbf{v}=\texttt{relaxvec}(\textbf{X}_1, \textbf{X}_2,\cdots, \textbf{X}_n,m,w);$
\STATE $\textbf{G}=\texttt{zeros}(w,w);$
\STATE \text{Compute} $\textbf{G}$ \text{and} $\widetilde{\textbf{G}}$ \text{defined as in} \eqref{e:GGt}.
\STATE $[\textbf{W},\textbf{D}]=\eig(\gamma\textbf{G}+(1-\gamma )\widetilde{\textbf{G}});$
\STATE $\textbf{W}=\textbf{W}(:,1:r);\textbf{D}=\diag(\textbf{D}(1:r)).$
\end{algorithmic}
\end{algorithm}

\begin{remark}
If there is no label information or people don't want to use it,  then let  $m=n$ (thus, $v_j=1/n,~n_j=1$) or $m=1$  (thus, $v_j=1,~n_j=n$),   $\widetilde{\textbf{G}}$ defined in \eqref{e:Gt} is exactly the total scatter matrix \eqref{e:G} defined for the classic 2DPCA just like the case $\gamma=1$.
\end{remark}

\begin{figure}[!h]
  \begin{center}
\includegraphics[height=0.3\textwidth,width=0.45\textwidth]{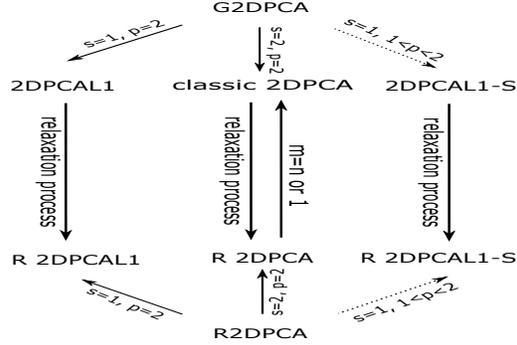}
  \end{center}
  \vskip-10pt \caption{\small relationship graph.}
\label{relationship}
\end{figure}

Now we focus on the relationships among  2DPCA, 2DPCA$L_1$, 2DPCA$L_1$-S , G2DPCA,  and R2DPCA.
It is obvious that 2DPCA and 2DPCA-$L_1$ are two special cases of G2DPCA.
2DPCA$L_1$-S originates from G2DPCA with $s=1$ and $p=1$ which leads to projection vector with only one nonzero element.
Then the $L_2$-norm constraint is employed to fix this problem, resulting in 2DPCA$L_1$-S.
On the other hand, G2DPCA with $s=1$ and $1<p<2$ behaves like 2DPCA$L_1$-S,  since the $L_p$-norm constraint in G2DPCA behaves like the mixed-norm constraint in 2DPCA$L_1$-S. Applying the relaxation idea to 2DPCA, 2DPCA$L_1$ and 2DPCA$L_1$-S, we can get three special cases of R2DPCA.
 To get a better understanding of these relationships, we construct a relationship graph in Fig \ref{relationship}.

\section{Experiments }\label{s:exs}
\noindent
In this section, we present numerical experiments  to compare the proposed relaxed two dimensional principle component analysis (R2DPCA) by $L_p$-norm   with  state-of-the-art algorithms on face recognition.
Three famous databases are utilized:
\begin{itemize}
\item Faces95 database (1440 images from 72 subjects, twenty images per subject),
\item color Feret database (3025 images from 275 subjects, eleven images per subject),
\item grey Feret database  (1400 images from 200 subjects, seven images per subject).
\end{itemize}
All of face images are cropped and resized,  and each image is  of 80$\times$80 size.
The numerical experiments are performed with MATLAB-R2016 on a personal computer with Intel(R) Xeon(R) CPU E5-2630 v3 @  2.4GHz (dual processor) and RAM 32GB.

\begin{example}\label{example1}
In this experiment, we compare R2DPCA with  2DPCA, 2DPCA-$L_1$, 2DPCA$L_1$-S, and  G2DPCA. 
We randomly select $10$ and $5$ images of each person from  Faces95 database and color Feret face database as the training set, respectively, and the remaining as the testing set.   As in \cite{wangj16}  we set  $\Omega=\{(s,p)|s=1.0:0.1:3.0, p=0.9:0.1:3.0\}$ for G2DPCA and  R2DPCA.  The parameter $\rho$ of 2DPCA$L_1$-S  relates to the $\lambda$ in \cite{zht06} via $\lambda=10^{-\rho}$ is tuned, consistent with \cite{whwj13}.  The optimal $\rho$ value is selected from $[-3.0:0.1:3.0]$. Here,  the relaxed parameter of criterion in \eqref{e:gtsc} is set as $\gamma=0$ and  the number of eigenfaces is fixed as  $r=10$  (Other cases will be considered in  Examples \ref{ex:gamma}-\ref{ex:eigenfaces}).   

We repeat the whole procedure  two times and output the average recognition rate. 
The face recognition rate (Accuracy) and corresponding optimal parameter are listed in Table \ref{faces95table} and Table \ref{THUMBNAILStable},  respectively. The reasonable trend of the classification accuracies according to different choices of $s$ and $p$ is presented in Fig \ref{faces95_sp} and Fig \ref{thum_sp}.
These numerical results indicate  that  R2DPCA  performs better than other four  state-of-art algorithms.

\begin{table}[!htbp]
\centering
\small{\caption{Classification accuracies of five algorithms on  faces95}{\label{faces95table}}}
\begin{tabular}{ccccccc}
 \hline
    $Algorithms$~~~~&  ~~~~$Optimal parameters$~~~~  &  ~~~~$Accuracy$~~~~  &\\ \hline
2DPCA&   $-$ & $0.8729$ \\
2DPCA-$L_1$&   $-$ &$0.8708$    \\
2DPCA$L_1$-S&  $\rho=-0.5$ &$0.8785$ \\
G2DPCA&   $s=2.7,p=2.2$ &$0.9451$    \\
R2DPCA $(\gamma=0)$&   $s=1,p=2.2$ &$\textbf{0.9493}$  \\\hline
\end{tabular}
\small
\end{table}
\begin{table}[!htbp]
\centering
\small{\caption{Classification accuracies of five algorithms on Color Feret}{\label{THUMBNAILStable}}}
\begin{tabular}{ccccccc}
 \hline
  $Algorithms$~~~~&  ~~~~$Optimal parameters$~~~~  &  ~~~~$Accuracy$~~~~  &\\ \hline
2DPCA&   $-$ & $0.5982$ \\
2DPCA-$L_1$&   $-$ &$0.5985$    \\
2DPCA$L_1$-S&  $\rho=-0.3$ &$0.6236$ \\
G2DPCA&   $s=2.8,p=2.6$ &$0.6918$    \\
R2DPCA $(\gamma=0)$&   $s=3,p=2.2$ &$\textbf{0.7085}$  \\\hline
\end{tabular}
\small
\end{table}

\begin{figure}[!h]
  \begin{center}
\includegraphics[height=0.30\textwidth,width=0.45\textwidth]{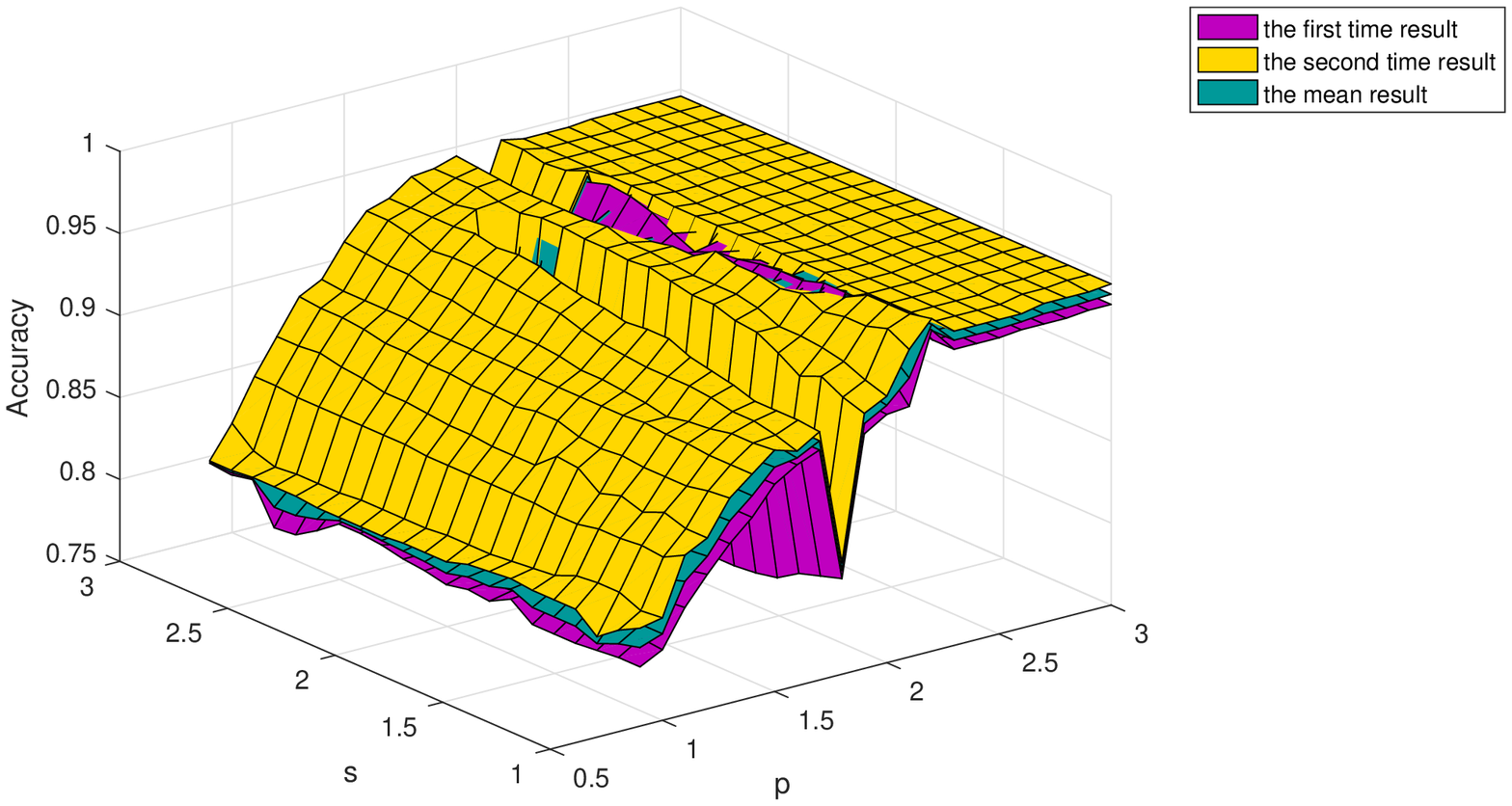}
  \end{center}
 \vskip-15pt \caption{\small Classification accuracies with $s, p$ on Faces95.}
\label{faces95_sp}
\end{figure}

\begin{figure}[!h]
  \begin{center}
\includegraphics[height=0.30\textwidth,width=0.45\textwidth]{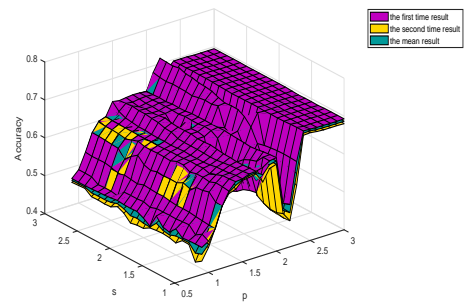}
  \end{center}
 \vskip-15pt \caption{\small Classification accuracies with $s, p$ on Color Feret.}
\label{thum_sp}
\end{figure}
\end{example}

\begin{example}\label{ex:gamma}
In this experiment, we research the effect of the parameter $\gamma$ of R2DPCA on the classification accuracy. 
The first $10$ and $5$ images of each person are selected  as the training sets of the Faces95  and  color/gray Feret  face databases,  respectively;  and $r=10$ features are selected.

The results with several representative values $\gamma$ are shown in Table \ref{t:gamma_omg} and Table \ref{t:gferet}. We can see that the Faces95 and color FERET databases are not sensitive to $\gamma$, and  however, we can see the validity of the parameters $\gamma$ on the Gray Feret database.  
\begin{table}[!htbp]
\centering
\caption{Classification accuracies according to different $\gamma$ on faces95.}
\label{t:gamma_omg}
\begin{tabular}{c|ccccccc}
 \hline
  Database&  $\gamma$~~~~&  ~~~~Optimal parameters~~~~  &  ~~~~Accuracy~~~~  &\\ \hline
faces95 &  $0:1/4:1$&   $s=1.1, p=2.2$ & ${\bf 0.8861}$\\
\hline
color FERET& $0:1/4:3/4$&   $s=3, p=2.2$ & $0.7673$ \\
&$1$&   $s=3,p=2.2$ &${\bf 0.7733}$   \\ \hline
\end{tabular}
\end{table}
\begin{table}[!htbp]
\centering
\caption{Classification accuracies according to  different $\gamma$ on Gray Feret.}
\label{t:gferet}
\begin{tabular}{cccccccccc}
 \hline
 $\gamma$~~~~&  ~~~~Optimal parameters~~~~  &  ~~~~Accuracy~~~~  &\\ \hline
    $0$&   $s=1.8, p=1.7$ & $0.6075$ \\
   $1/4$&   $s=1.9, p=1.7$ & $0.6075$ \\
    $1/2$&   $s=2.3, p=1.6$ & ${\bf 0.6112}$ \\
   $3/4$&   $s=2.4, p=1.6$ & $0.6088$ \\
   $1$ &   $s=2.6,p=1.8$ & $0.5837$  \\
\hline
\end{tabular}
\end{table}
\end{example}

\begin{example}\label{ex:eigenfaces}
In this experiment, we test the effect of numbers of chosen features on  the classification accuracy. 
We randomly select $10, 5$ images of each subject as training samples and the remaining as testing samples on the Faces95 and Color Feret databases, respectively. The whole procedure is  repeated two times and  the average accuracies are listed.  Based on  the optimal parameters $s, p$ of R2DPCA with $\gamma=0$ in Example \ref{example1}, we set  $s=1,~~p=2.2$ and $s=3,~~p=2.2$.

Fig \ref{faces95_experiment2} and Fig \ref{thum_experiment2} show the classification accuracies of G2DPCA and R2DPCA with different feature numbers in the range of $[1,30]$ on the Faces95 database and Color Feret database, respectively. From these results, we can see that the classification accuracies of R2DPCA are higher and more stable than G2DPCA. 
When $k=1$ the classification accuracies of G2DPCA and R2DPCA are the same, which consists to the theory. 
\begin{figure}[!h]
  \begin{center}
\includegraphics[height=0.30\textwidth,width=0.45\textwidth]{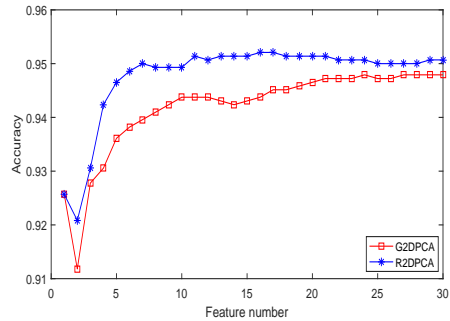}
  \end{center}
  \vskip-15pt \caption{\small Classification accuracies of R2DPCA and G2DPCA with $k=[1:30]$ on Faces95.}
\label{faces95_experiment2}
\end{figure}
\begin{figure}[!h]
  \begin{center}
\includegraphics[height=0.30\textwidth,width=0.45\textwidth]{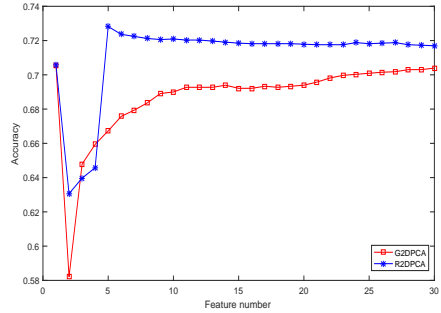}
  \end{center}
  \vskip-15pt \caption{\small Classification accuracies of R2DPCA and G2DPCA with $k=[1:30]$ on Color Feret.}
\label{thum_experiment2}
\end{figure}
\end{example}

\begin{example}\label{example3}
In this experiment, we research the influence of parameters $s$, $p$ on classification accuracies of R2DPCA with the case $\gamma=0$.

  The training sets and testing sets just like Example \ref{example1}.
  The classification accuracies with $10$ feature numbers, then the results are recorded. The procedure is repeated two times and then we take the average value. We use the optimal parameters of each databases on R2DPCA in Example \ref{example1}. Here we fix $s=1$ and search the optimal parameters set from $p=[0.9:0.1:3.0]$ on the Faces95 database. We fix $s=3$ and search the optimal parameters set from $p=[0.9:0.1:3.0]$ on the Color Feret database. Similarly, we fix $p=2.2$ and search the optimal parameters set from $s=[1.0:0.1:3.0]$ on the Faces95 database. We fix $p=2.2$ and search the optimal parameters set from $s=[1.0:0.1:3.0]$ on the Color Feret database.

\begin{figure}[!h]
  \begin{center}
\includegraphics[height=0.30\textwidth,width=0.45\textwidth]{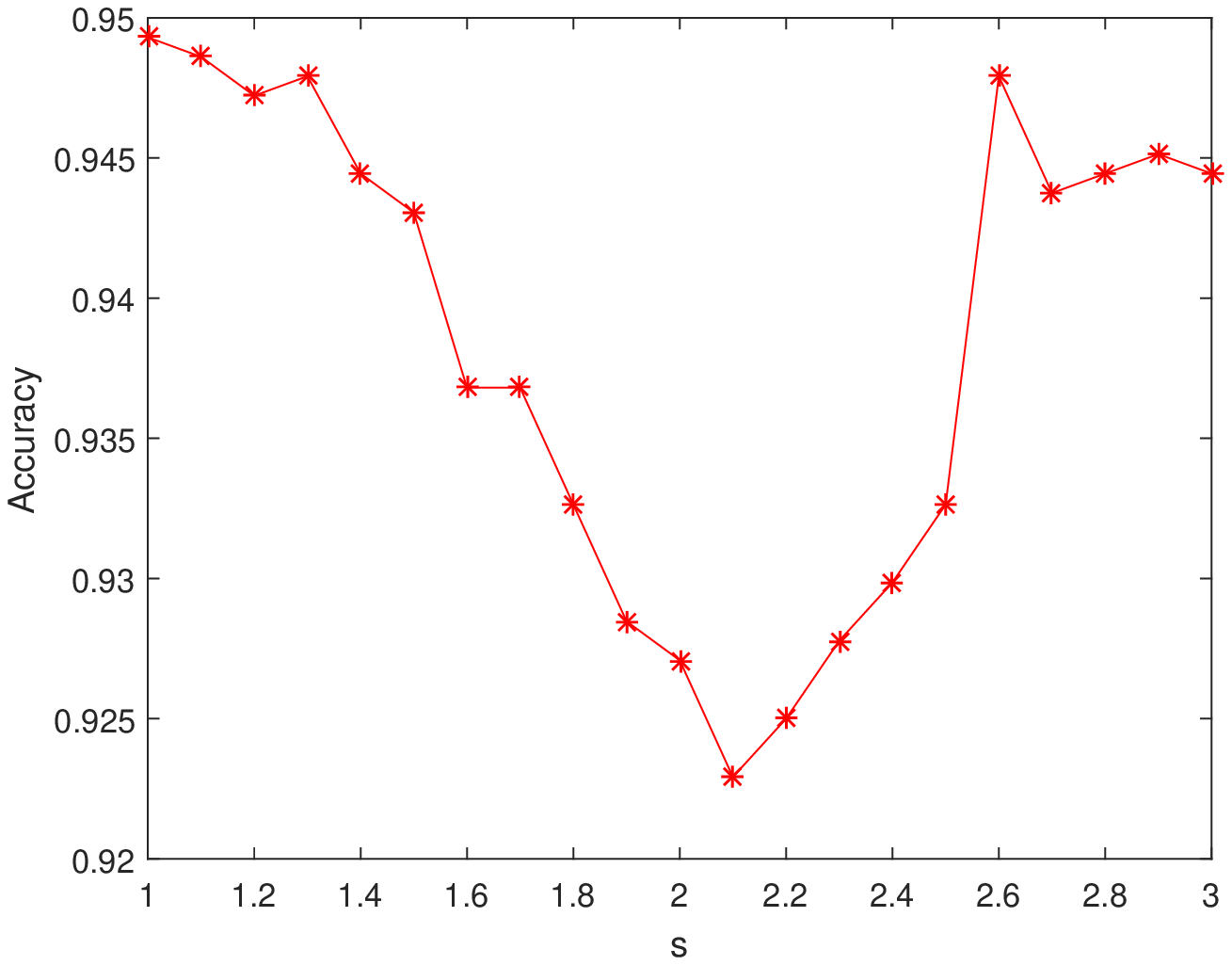}
\includegraphics[height=0.30\textwidth,width=0.45\textwidth]{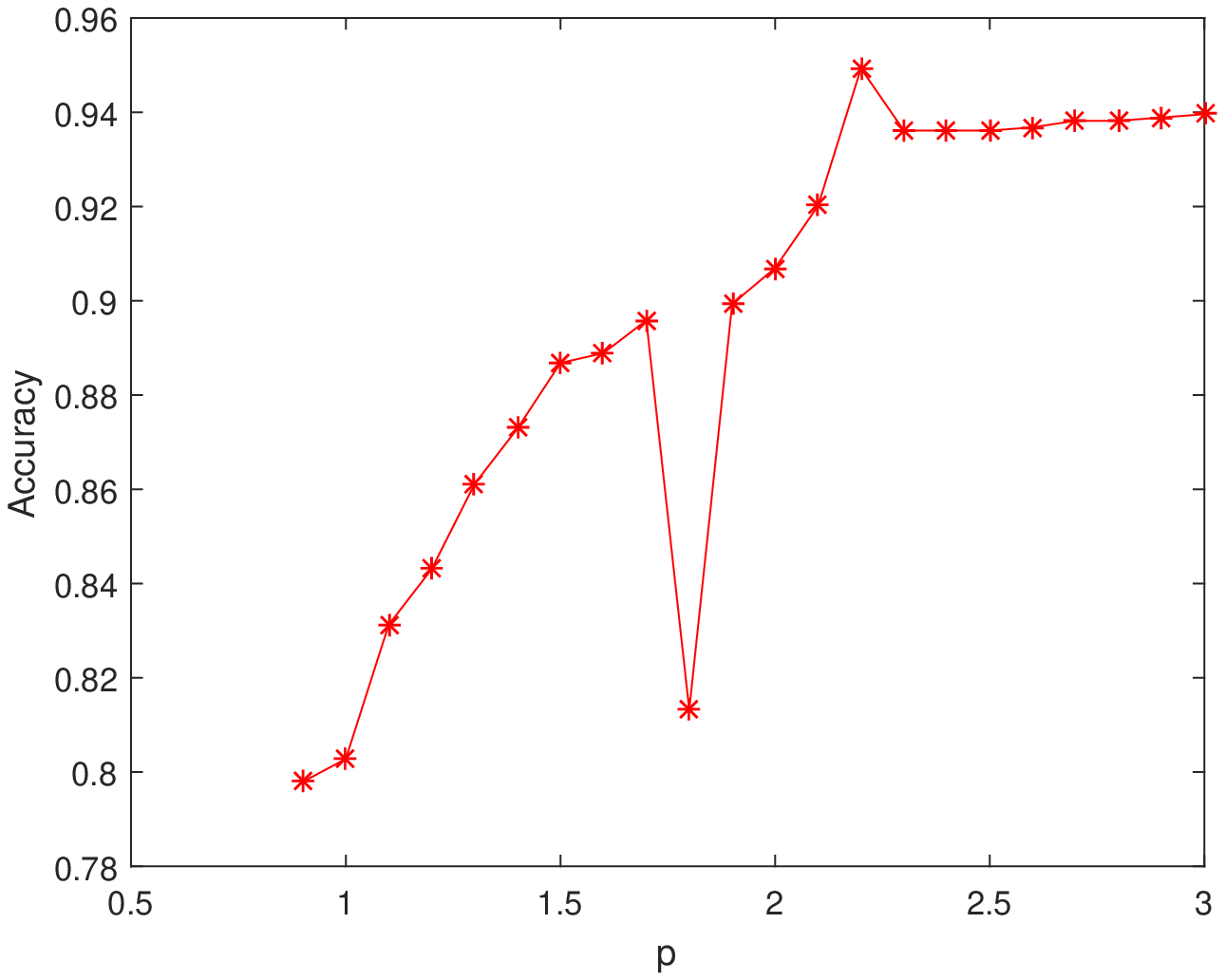}
  \end{center}
  \vskip-15pt \caption{\small Classification accuracies of R2DPCA with $s$ and $p$ on Faces95.}
\label{faces95_experiment3}
\end{figure}

\begin{figure}[!h]
  \begin{center}
\includegraphics[height=0.30\textwidth,width=0.45\textwidth]{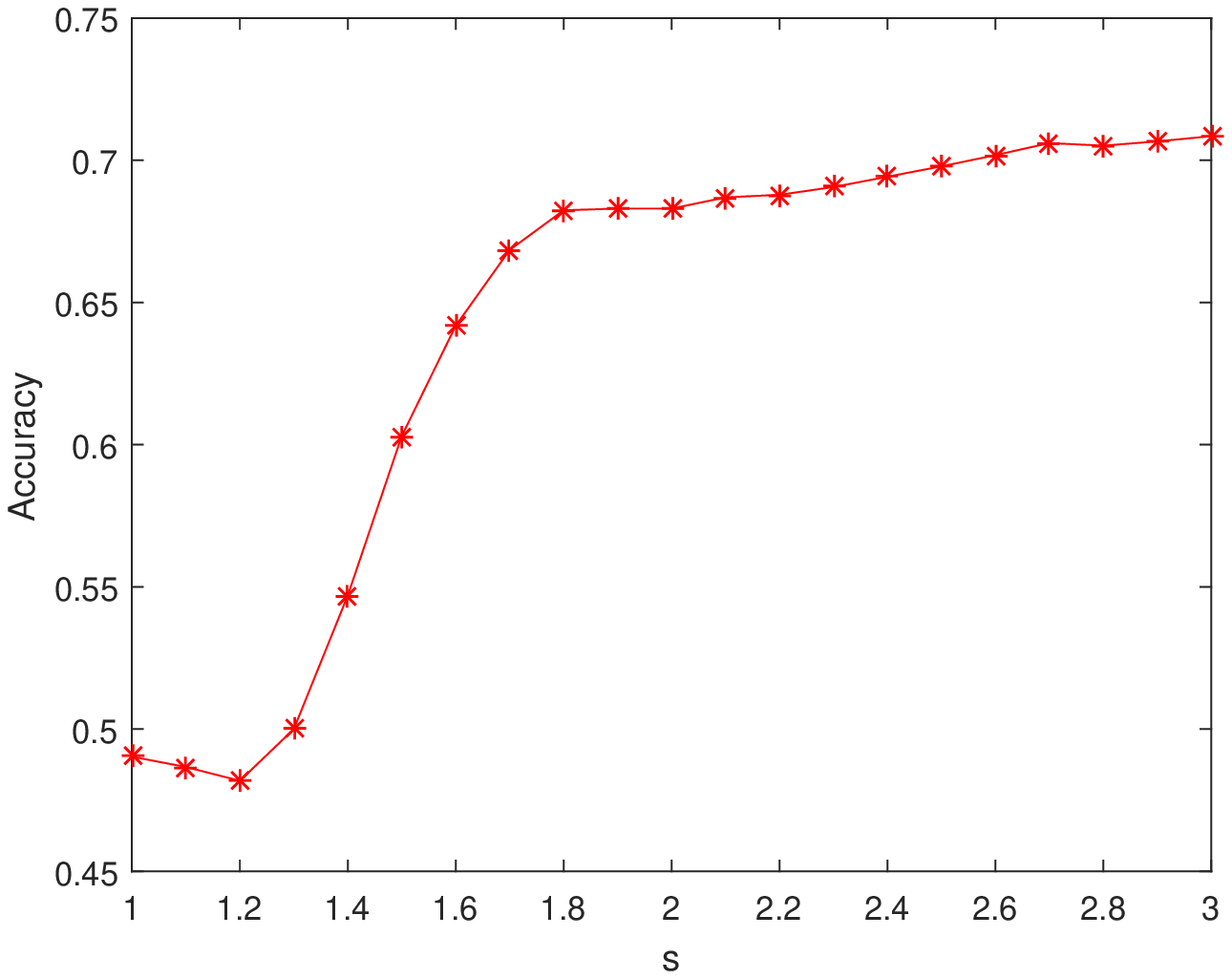}
\includegraphics[height=0.30\textwidth,width=0.45\textwidth]{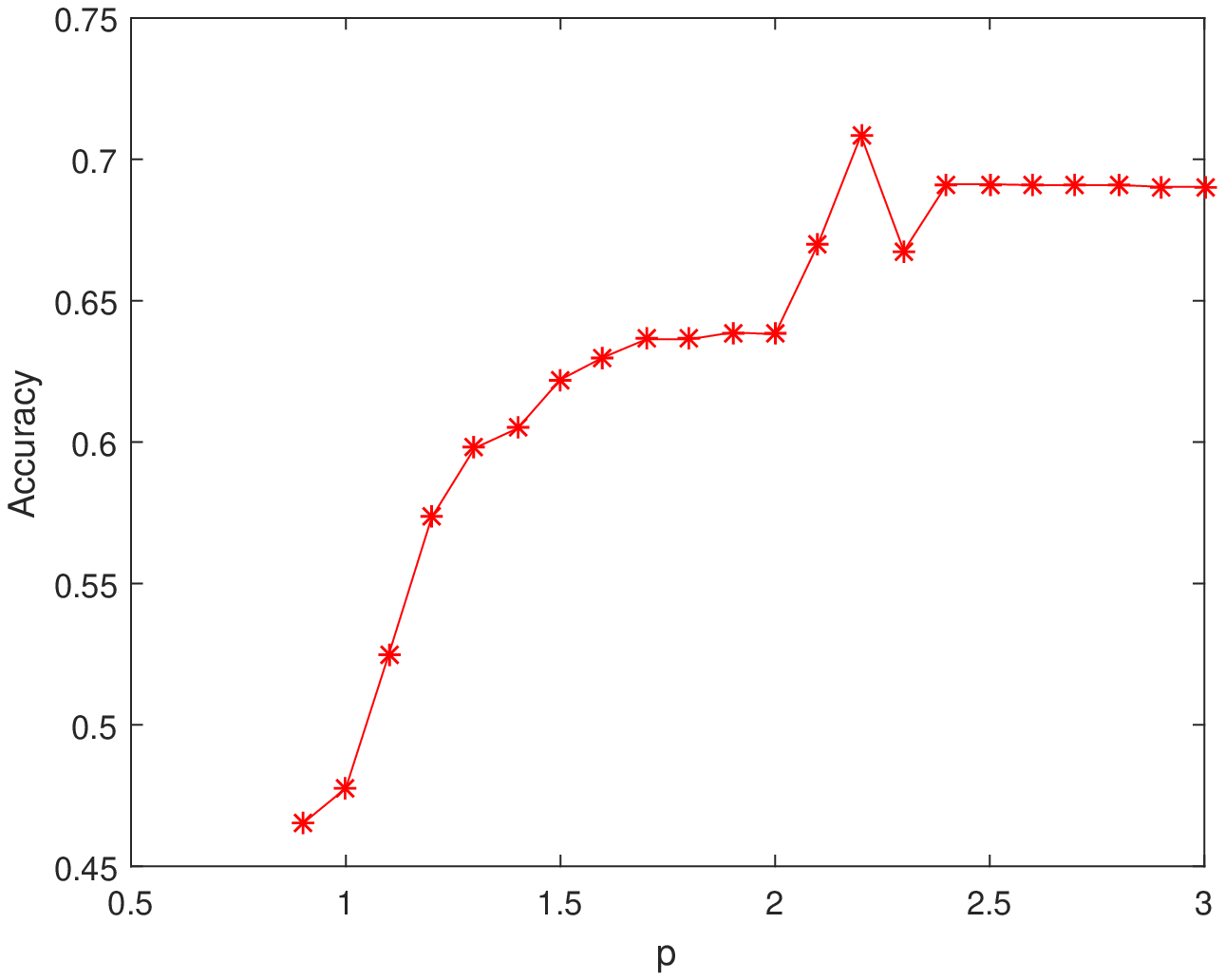}
  \end{center}
  \vskip-15pt \caption{\small Classification accuracies of R2DPCA with $s$ and $p$ on Color Feret.}
\label{thum_experiment3}
\end{figure}

The results are shown in Fig \ref{faces95_experiment3} and Fig \ref{thum_experiment3}. From these Figures, we know that when $s=1, p=2.2$,  the accuracy classification approach to maximum on the Faces95 database. When $s=3, p=2.2$, the accuracy classification approach to maximum on the Color Feret database. These results are consistent with the results of Experiment \ref{example1}. And from these figures, we also can know that the accuracy classification do not have a stable variation trend with $s$ or $p$.
\end{example}

\begin{example}
   In section \ref{ss:radsm}, we proposed a restarted alternating direction search method. Now we test this method on the Faces95 database and Color Feret database. From Example \ref{example3}, we know the classification accuracies don't have a stable trend with different $s$ or $p$. Traditionally, we need to traverse all the combinations of $s, p$, so that we can get the maximum solution. This way spends large time of calculations. Here, we test the restarted alternating direction search method.

    $\Omega=\{(s,p)|s=1.0:0.1:3.0, p=0.9:, 0.1:3.0\}$ is set  as in \cite{wangj16}. We randomly use $10, 5$ images of each subject as the training samples and the remaining images as the testing samples to do this restarted alternating direction search algorithm. We randomly start from four initial points, i.e., $s=1, s=1.6, p=0.9, p=1.2 $ to find corresponding starters and then find the max classification accuracy of R2DPCA. We set a value $s$ (or $p$) to find a value $p$ (or $s$) and use $(s, p)$ as a starter. We set $\delta=0.3$ to control time. We do R2DPCA with $\gamma=0$ and obtain the max classification accuracy $0.9375, 0.7158$ on two databases respectively. In order to have a more intuitive view of searching paths, we describe the results in Table \ref{faces95_searching_path} and Table \ref{thum_searching_path}.

   From section \ref{ss:radsm}, if $\varrho^1(i,j)=\varrho^0(i,j)$ and  $(s_i^1,p_j^1)=(s_i^0,p_j^0)$, we should go to Step 5 to do a restarted algorithm. This situation occurs in this experiment. For example, in table \ref{faces95_searching_path}, we can see that if starter is $s=1, p=2.2$, we  make a restarted algorithm in the next step. Because of a small positive value $\delta=0.3$, it can't reach to the new starter correspond to the next max accuracy which can be seen from the second row in this table. If we set a enough bigger positive value $\delta$, it can search the optimal solution. Due to this property, we can use this algorithm to do a pre computation with databases which needed to be identified, so that we can have a general idea of whether these databases are applicable to the algorithms we proposed.

    In each two databases, we should do Algorithm \ref{a:r2dpca} 462 times in traditional R2DPCA. But now we at most do Algorithm \ref{a:r2dpca} 153 times and even do 3 times at least. And there is a little difference between observed results and true accuracies.

 \begin{table*}[!htbp]
\centering
\tiny{\caption{Restarted alternating searching path on Face95.}{\label{faces95_searching_path}}}
\begin{tabular}{ccccccc}
 \hline
    $initial ~~~~ point$~~~~& $path$~~~~  &  ~~~~$searching~~~~  Accuracy$~~~~  &\\ \hline
$s=1, p=2.2$&   $(1, 2.2)$ & $0.9347$ \\
$s=1.3, p=0.9$&   $(1.3, 0.9) \rightarrow (1.3, 2.1) \rightarrow (1.8, 2.1)$ &$\textbf{0.9375}$    \\
$s=1.6, p=2.8$&  $(1.6, 2.8) \rightarrow (1.3, 2.8) $&$0.9306$\\
$s=2.7, p=1.2$&   $(2.7,1.2)\rightarrow (2.7,2.2)\rightarrow (1,2.2)\rightarrow (1,2.1)\rightarrow (1.8, 2.1)$ &$\textbf{0.9375}$    \\\hline
\end{tabular}
\small
\end{table*}
  \begin{table*}[!htbp]
\centering
\tiny{\caption{Restarted alternating searching path on Color Feret. }{\label{thum_searching_path}}}
\begin{tabular}{cccccc}
 \hline
    $initial ~~~  point$~~~~&  $path$~~~~  &  ~~~~$searching~~~~  Accuracy$~~~~  &\\ \hline
$s=1, p=2.4$&   $(1, 2.4) \rightarrow (2.7, 2.4)\rightarrow (2.7, 2.2)$ & $\textbf{0.7158}$ \\
$s=1.7, p=0.9$&   $(1.7, 0.9) \rightarrow (1.7, 2.3) \rightarrow (1.6, 2.3)\rightarrow (1.6,2.6)\rightarrow (2.5,2.6)\rightarrow (2.5,2.1)$ &$0.7085$    \\
$s=1.6, p=1.7$&  $(1.6, 1.7) \rightarrow (2.5, 2.7) \rightarrow (2.5, 2.2) \rightarrow (2.5, 2.2)\rightarrow(2.7,2.2) $ &$\textbf{0.7158}$ \\
$s=1.2, p=1.2$&   $(1.2,1.2)\rightarrow (1.2,2.3)$ &$0.6939$    \\\hline
\end{tabular}
\small
\end{table*}

\end{example}

\begin{example}
This method is common to 2DPCA-like methods. Now we compare the results of algorithms including 2DPCA, 2DPCA-$L_1$, 2DPCA$L_1$-S with their relaxation results on Gray Feret database. 

We randomly select $tr=3$ training samples from each subject and the remaining images as testing samples. Here, we also choose 10 feature numbers to save computational time. Then the nearest neighbour classifier is applied to do classification. Also, the procedure is repeated two times and take the average classification accuracies. Notice that the relaxed 2DPCA Algorithm \ref{relaxed_2dpca} as shown in Remark \ref{remark1} and the other two algorithms' relaxed progression are similar with R2DPCA Algorithm \ref{a:r2dpca}. In order to be consistent with the previous experimental parameters, we set $\gamma=0$. The results see Table \ref{feret_example5}.

\begin{table*}[!htbp]
\centering
\small{\caption{classification accuracies of six algorithms on Gray FERET}{\label{feret_example5}}}
\begin{tabular}{ccccccc}
 \hline
    $Algorithms$~~~~&  ~~~~$Optimal parameters$~~~~  &  ~~~~$Accuracy$~~~~  &\\ \hline
 2DPCA&   $-$ &$0.4225$    \\
R2DPCA $(s=p=2)$&   $-$ &$\textbf{0.6387}$  \\\hline
2DPCA-$L_1$&   $-$ & $0.4225$ \\
R2DPCA$(s=1,p=2)$&   $-$ &$\textbf{0.6123}$    \\\hline
2DPCA$L_1$-S&  $\rho=-0.6$ &$0.4938$ \\
R2DPCA $(s=1)$&   $\rho=-2$ &$\textbf{0.5988}$    \\\hline
\end{tabular}
\small
\end{table*}
From the results, we can know that our proposed R2DPCA algorithm is also effective  for other 2-D algorithms.
\end{example}

\begin{example}
We test the accuracies of the Gray Feret database with the number of training samples in this example.

We randomly select $tr=3$ training samples from each subject and the remaining images as testing samples. Here, we also choose 10 feature numbers. Then the nearest neighbor classifier is applied to do classification. We do this process two times and take the average classification accuracies in Fig \ref{grayferet_database}. Also, in order to be consistent with the previous experimental parameters, we set $\gamma=0$.
\begin{figure}[!h]
  \begin{center}
\includegraphics[height=0.30\textwidth,width=0.45\textwidth]{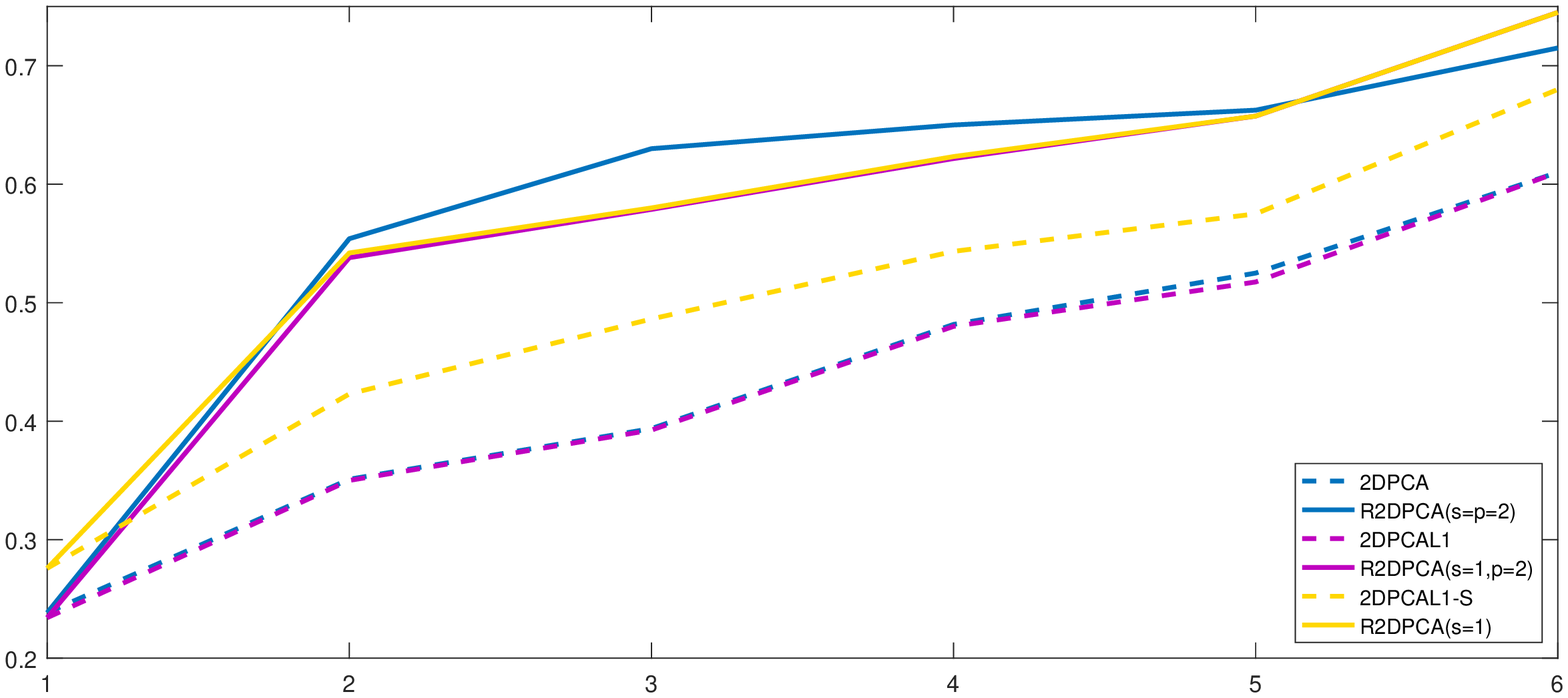}
  \end{center}
  \vskip-15pt \caption{\small Accuracy with the number of training samples on Gray Feret database.}
\label{grayferet_database}
\end{figure}

From this Figure, we known that the classification accuracies of relaxed versions are higher than those in original versions.
\end{example}

\section{Conclusion}\label{s:conclusion}
In this paper, we present a relaxed two dimensional  principal component analysis (R2DPCA) approach for face recognition, with applying the label information of the training data. The R2DPCA is a generalization of 2DPCA, 2DPCA-$L_1$ and G2DPCA, and has higher generalization ability. Since utilizing the label information, the R2DPCA can be seen as  a new supervised projection method, but it is totally different to  the two-dimensional linear discriminant analysis (2DLDA)\cite{ye05,lls08}.

\section*{Acknowledgments}
This paper is supported in part  by  National Natural Science
 Foundation of China  under grants 11771188 and 
 a Project Funded by the Priority Academic Program Development
  of Jiangsu Higher Education Institutions.


\end{document}